\pdfoutput=1
\documentclass[11pt]{article}%
\usepackage{amsfonts}
\usepackage{amssymb}
\usepackage{graphicx}
\usepackage{setspace}
\usepackage{amsmath}%
\usepackage{amsthm}%
\usepackage{bbm}
\usepackage{palatino}
\usepackage{paralist}
\usepackage{multirow}
\usepackage{booktabs}
\usepackage{dsfont}
\usepackage{indentfirst}
\usepackage{enumerate}
\usepackage{enumitem}
\usepackage{longtable}
\usepackage{bm}
\usepackage{caption}
\usepackage{subcaption}
\usepackage{mwe}
\usepackage{float}
\usepackage{booktabs}
\usepackage{lscape}
\usepackage[hyperindex,breaklinks]{hyperref}
\hypersetup{colorlinks=true,       % false: boxed links; true: colored links
	linkcolor=red,       
	citecolor=blue,        % color of links to bibliography
	filecolor=magenta,      % color of file links
	urlcolor=cyan           % color of external links
}
\usepackage{xskak}

\usepackage{xcolor,csquotes,bm}
\usepackage{tikz}

\usetikzlibrary{shapes.misc}

\theoremstyle{definition}

\DeclareMathOperator*{\argmax}{\arg\!\max}
\newcommand{\norm}[1]{\left\lVert#1\right\rVert}

\usepackage[backend=biber, style=authoryear, eprint=false, doi=false, isbn=false, date=year, natbib]{biblatex}
\bibliography{studies_paper}

\title{Chess AI: Competing Paradigms for Machine Intelligence}

\author{		
    \makebox[.2\linewidth]{Shiva Maharaj\thanks{Email: \href{mailto:problemsolver2020@gmail.com}{problemsolver2020@gmail.com}}}\\
         \textit{\small  ChessEd}
	\and
    \makebox[.2\linewidth]{Nick Polson\thanks{Email: \href{mailto:ngp@chicagobooth.edu}{ngp@chicagobooth.edu}}}\\
	\textit{\small  Booth School of Business}\\
	\textit{\small  University of Chicago}\\
        \and
        \makebox[.2\linewidth]{Alex Turk\thanks{Email: \href{mailto:alexwturk@gmail.com}{alexwturk@gmail.com}}}\\
	\textit{\small Phillips Academy}
}

\graphicspath{{../fig/}{../wine/}}

\begin{document}
\maketitle

%%%%%%%%%%%%%%%%%%%%%

\begin{abstract}
\noindent  
Endgame studies have long served as a tool for testing human creativity and intelligence. We find that they can serve as a tool for testing machine ability as well. Two of the leading chess engines, Stockfish and Leela Chess Zero (LCZero), employ significantly different methods during play. We use Plaskett's Puzzle, a famous endgame study from the late 1970s, to compare the two engines. Our experiments show that Stockfish outperforms LCZero on the puzzle. We examine the algorithmic differences between the engines and use our observations as a basis for carefully interpreting the test results. Drawing inspiration from how humans solve chess problems, we ask whether machines can possess a form of imagination. On the theoretical side, we describe how Bellman's equation may be applied to optimize the probability of winning. To conclude, we discuss the implications of our work on artificial intelligence (AI) and artificial general intelligence (AGI), suggesting possible avenues for future research.

\end{abstract}

\newpage

\section{Introduction}
\vspace{0.1in} 

\emph{Chess is not a game. Chess is a well-defined form of computation. You may not be able to work out the answers, but in theory, there must be
a solution, a right procedure in any position---John von Neumann}

\vspace{0.1in} 

There are various schools of thought in chess composition, each of which place different emphasis on the complexity of problems. Chess studies originally became popular in the 19th century. They are a notoriously difficult kind of puzzle, involving detailed calculations and tactical motifs. As such, they provide a good benchmark for AI studies.

We choose to analyze how chess engines respond to Plaskett's Puzzle, one of the most well-known endgame studies in history. There are many features of this problem that make it particularly hard---not just the depth required ($15$ moves) but also the use of underpromotion and subtle tactical strategies. Though the puzzle was initially created by a Dutch composer, it achieved notoriety in 1987 when English grandmaster Jim Plaskett posed the problem at a chess tournament, stumping all players except Mikhail Tal.

The advent of AI algorithms and powerful computer chess engines enables us to revisit chess studies where the highest level of tactics and accuracy are required.
We evaluate the performance of Stockfish 14 \citep{romstad_stockfish_2021} and Leela Chess Zero \citep{pascutto_leela_2018} on Plaskett's Puzzle. We discuss the implications of their performance for both end users and the artificial intelligence community as a whole. We find that Stockfish solves the puzzle with much greater efficiency than LCZero. Our work implies that building a chess engine with a broad and efficient search may still be the most robust approach.

Artificial general intelligence (AGI) is the ability of an algorithm to achieve human intelligence in a multitude of tasks.  \citet{feynman_pleasure_1981} famously claimed that the rules of chess could be determined from empirical observation, a central tenet of modern machine intelligence.  Historical records provide insight into human performance on Plaskett's Puzzle. By comparing this with machine performance, we can better assess the state of current progress on the long road to building AGI. We further discuss which algorithms possess greater potential in the field of AGI by reviewing their ability to generalize to different domains.

On the theoretical side, we describe a central tool for solving dynamic stochastic programming problems: the Bellman equation. This framework, together with the use of deep neural networks to model the value and policy functions, provides a solution for maximising the probability of winning the chess game. In doing so, the algorithm offers an optimal path of play. Given the enormous number of possible paths of play (the Shannon number), the use of search methods and the depth of the algorithm become important computational aspects.  It is here that high level chess studies provide an important mechanism for testing the ability of a given AI algorithm.

The rest of the paper is outlined as follows. The next subsection provides historical perspective on Plaskett's Puzzle.
Section 2 provides background material on current AI algorithms used in chess, namely, Stockfish 14 and LCZero. Section 3 provides our detailed analysis of   Plaskett's Puzzle using these chess engines. Finally,  Section 4 concludes with directions for future research.  In particular, we discuss the implications of our work for human-AI interaction in chess \citep{polson_aiq_2018} and AI development. 

\subsection{Plaskett's study} 

The story of Plaskett's puzzle dates to a Brussels tournament in 1987. 
The study was originally composed by Gijs van Breukelen in 1970. In 1987, it famously stumped multiple super grandmasters when presented by James Plaskett in a tournament press room. It was finally solved that day by legendary attacking player Mikhail Tal, who figured it out during a break at the park. \citep{friedel_solution_2018}

The highly inventive puzzle involves multiple underpromotions and was originally designed to be a checkmate in 14. There is a mistake in the original puzzle whereby black can escape mate, though white is still winning in the final position. This mistake can be corrected by moving the black knight on g5 to h8. Harold van der Heijden's famous {\em Endgame Study Database} (which contains over $58,000$ studies) proposes another corrected version where he moves the black knight from g5 to e5. 
Figure 1 shows Plaskett's original puzzle.

\begin{center}
\fenboard{8/3P3k/n2K3p/2p3n1/1b4N1/2p1p1P1/8/3B4 w - - 0 1}
\showboard \\
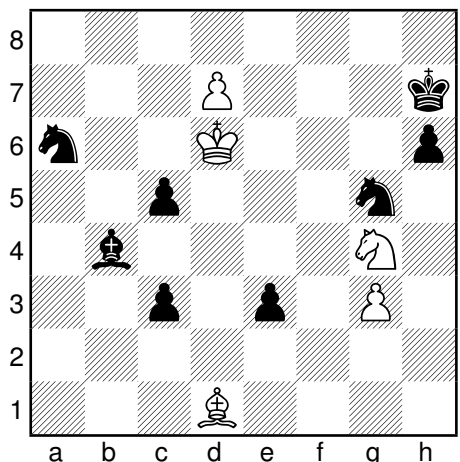
\emph{Figure 1: Initial board position}
\end{center}

\section{Chess AI}

\citet{turing_chess_1953}, von Neumann, and \citet{shannon_programming_1950} pioneered the  development of AI algorithms for solving chess. 
The Shannon number measures the number of possible states of a system. For chess, this number is $ 10^{152} $, making the game a daunting computational challenge. A major advance over pure look-ahead search methods was the use of deep neural networks to approximate the value and policy functions.  Then, the self-play of the algorithms allows for quick iterative solution paths. See \citet{silver_mastering_2017} for further discussion. 

The dynamic programming method breaks the decision problem into smaller sub-problems. Bellman's principle of optimality describes how to do this:

\vspace{0.1in}

\emph{Bellman Principle of Optimality: An optimal policy has the property that whatever the initial state and initial decision are, the remaining decisions must constitute an optimal policy with regard to the state resulting from the first decision. \citep{bellman_dynamic_1957}}

\vspace{0.1in}

Backwards Induction identifies what action would be most optimal at the last node in the decision tree (a.k.a. checkmate). Using this information, one can then determine what to do at the second-to-last time of decision. This process continues backwards until one has determined the best action for every possible situation (a.k.a. solving the Bellman equation).

\subsection{Q-values} 

The optimal sequential decision problem is solved by calculating the values of the $Q$-matrix, denoted by $Q(s,a)$ for state $s$ and action $a$. One iterative process for finding these values is known as $Q$-learning \citep{watkins_q-learning_1992}, which can be converted into a simulation algorithm as shown in \citet{polson_simulation-based_2011}.
The $Q$-value matrix describes the value of performing action $a$ (a chess move) in our current state $s$ (the chess board position) and then acting optimally henceforth.
The current optimal policy and value function are given by 
\begin{align*}
V(s) & = \max_a \; Q ( s , a )  = Q( s , a^\star (s)  )  \\
 a^\star (s) & = {\rm arg max}_a \;   Q ( s , a )
\end{align*} 
For example, chess engines such as LCZero simply take the probability of winning as the objective function. Hence, at each stage $V(s)$ measures the probability of winning.
This is typically reported as a centi-pawn advantage. 

The Bellman equation for $Q$-values (assuming instantaneous utility $u(s,a)$ and a time inhomogeneous $Q$ matrix), is the constraint 
$$
Q( s , a) = u(s,a) +  \sum_{ s^\star \in S } P( s^\star | s ,a ) \max_{ a } Q ( s^\star , a )
$$
Here $P (s^\star | s ,a )$ denotes the transition matrix of states and describes the probability of moving to new state $ s^\star $ given current state $s$ and action $a$.
The new state is clearly dependent on the current action in chess and not a random assignment. 
Bellman's optimality principle is therefore simply describing the constraint for optimal play as one in which the current value is a sum over all future paths of the 
probabilistically weighted optimal future values of the next state.

Taking maximum value of $ Q(s,a)$ over the current action $a$ yields 
\begin{align*}
V(s)   =  \max_a \left \{ u(s,a) +  \sum_{ s^\star \in S } P( s^\star | s ,a )   V (s^\star) \right \}  \; \; {\rm where} \; \;  V (s^\star) & = {\rm max}_a \;   Q ( s^\star , a ) .
\end{align*} 
Here $u(s,a)$ is an instantaneous utility obtained from action $a$ in state $s$. At the end of the game, checkmate will lead to $u(s,a)=1$.

\subsection{Stockfish 14 anatomy}

\subsubsection{Search}
Stockfish uses the alpha-beta pruning search algorithm \citep{edwards_alpha-beta_1961}. Alpha-beta pruning improves minimax search \citep{wiener_information_1948, von_neumann_zur_1928} by avoiding variations that will never be reached in optimal play because either player will redirect the game.

Since it is often computationally infeasible to search until the end of the game, the search is terminated early when it reaches a certain depth. Search depth is measured by \textit{ply}, where a ply is a turn taken by a player. A search depth of $D$ indicates that the distance between the root node and the leaf nodes of the search tree is $D$ ply. Stockfish incrementally increases the depth of its search tree in a process known as \emph{iterative deepening} \citep{groot_thought_1978}. However, when a nominal search depth of $D$ is reported by chess engines, it does not mean that the search has considered all possible variations of $D$ moves. This is due to heuristics which cause the engine to search promising variations to a greater depth than nominal and less promising variations to a lesser depth than nominal.

The engine applies two main classes of heuristics to reduce the search space: forward pruning and reduction \citep{isenberg_stockfish_2021}. Forward pruning techniques remove game tree subgraphs that are unlikely to be contained in optimal play. For example, if the evaluation of a position is significantly worse than the value guaranteed by a player's best alternative, the position's children are pruned early. This is known as futility pruning \citep{schaeffer_experiments_1986, heinz_extended_1998}. It is possible that the engine mistakenly prunes a line of play. This will be corrected once the engine depth surpasses a technique-specific depth cap\footnote{See Stockfish source code at \url{https://github.com/official-stockfish/Stockfish}}, after which the technique is no longer applied.

Reduction techniques search certain game tree subgraphs to lower depths, rather than omitting them from the search altogether. A canonical example is late move reductions \citep{levy_sex_1989}, which assume that the engine checks better moves earlier. Moves checked later are searched to lower depths than nominal.

Given infinite time, the engine will converge to the optimal line of play. Depth caps prevent lines from being overlooked via pruning, and reductions become inconsequential at infinite depth.

\subsubsection{Evaluation}
Once the search algorithm reaches a leaf node, a heuristic evaluation function is applied to determine whether the ending position favors White or Black. In Stockfish versions 11 and under, this function is hard-coded based on chess concepts such as piece position, piece activity, and the game phase (opening/middle game or endgame).\footnote{\url{https://hxim.github.io/Stockfish-Evaluation-Guide/}}

The efficiently updatable neural network (NNUE) evaluation function was originally invented by \citet{nasu_nnue_2018} for Shogi, a Japanese chess variant. Stockfish implemented it for their chess engine in version 12 \citep{the_stockfish_team_nnue_2020}. The neural network is trained to predict the output of the classic Stockfish evaluation function at "moderate" search depths. Thus, it can be thought of as a search depth multiplier. Though NNUE is about twice as slow as Stockfish's classic evaluation function, the engine makes up for this in terms of evaluation quality \citep{isenberg_nnue_2021}.

Its architecture comprises a shallow, four layer neural network specifically optimized for speed on CPU machines. The input layer of binary features describes the position of the White and Black king relative to the White and Black pieces, respectively. For example, one feature might be described as

\[ \theta_0 =
   \begin{cases}
   1 & \text{if Black  \makebox[1em]{\symking}g8 and  \makebox[1em]{\symqueen}d8} \\
   0 & \text{otherwise}
   \end{cases}
\]

The dimensionality of the board feature vector is reduced as it passes through two hidden layers and one output layer, resulting in a final scalar value for the position evaluation.

\subsection{AlphaZero anatomy}

\subsubsection{Search}
AlphaZero uses the Monte Carlo Tree Search (MCTS) algorithm to identify the best lines via repeated sampling \citep{silver_general_2018}. In MCTS, node evaluations at the end of prior simulations are used to direct future simulations toward the most promising variations. The search depth increases with each simulation. After some number of simulations, the child node with the most samples is chosen. Node value is optimistically estimated by the Polynomial Upper Confidence Tree (PUCT) algorithm \citep{rosin_multi-armed_2011, silver_mastering_2017}:
\begin{align*}
    a_t = \argmax_a (Q(s_t, a) + U(s_t, a)) \\
    U(s, a) = C(s) P(s, a) \frac{\sqrt{\sum_b N(s, b)}}{1 + N(s, a)} \\
    C(s) = \log {\frac{1 + N(s) + c_{base}}{c_{base}}} + c_{init}
\end{align*}

Both the action value $Q(s, a)$ and policy $P(s, a)$ are determined by applying the evaluation function to a state in the game tree. The search algorithm initially focuses on nodes with high $Q$-value and prior probability $P$. As $N(s, a)$, the number of times action $a$ has been taken from state $s$, increases, the algorithm increasingly relies on the sampled $Q$-value. $C(s)$ controls the amount of exploration, which increases as the search progresses.

\subsubsection{Evaluation}
AlphaZero's evaluation uses deep convolutional neural networks (CNNs) to estimate the policy vector $\mathbf{p}_t$ and value $v_t$ of nodes in the search tree \citep{silver_general_2018}. Data is generated through millions of games of self-play. Given the final outcome of the game $z$ ($-1$ for loss, $0$ for draw, $+1$ for win) and the search probabilities $\pi_t$ (obtained from the final node visit counts), the networks are trained to minimize the loss of
\[
    (z - v)^2 - \pi^T \log{\mathbf{p}} + c \norm{\theta}^2
\]

While $v$ aims to directly predict the game result, $\mathbf{p}$ is trained to "look ahead" in the search, causing it to learn helpful prior probabilities for the PUCT algorithm.

Early neural chess engines often required hand-crafted feature representations that went beyond the board position and basic board statistics. For example, the NeuroChess \citep{thrun_learning_1995} input features included the number of weak pawns on the board and the relative position of the knight and queen. With the advent of deep learning, the process of performing feature selection by hand can be eliminated, as additional layers perform a sophisticated—and perhaps even more effective—form of feature selection \citep{lecun_deep_2015}. The learner relies on the "unreasonable effectiveness of data" to learn useful features from unstructured input \citep{halevy_unreasonable_2009}.

In particular, AlphaZero's CNN inputs are binary $8\times8$ feature planes that encode the piece locations of both players for the past 8 half-moves. In addition, the input includes constant feature planes which encode important counts in the game, such as the repetition and move count. Like the input, the output policy $\mathbf{p}_t = P(a \mid s_t)$ is represented in the network by a stack of planes. Each of the planes in the $8 \times 8 \times 73$ stack represents a movement-related modality, and each square represents the location from which to pick a piece up. For example, the first $8 \times 8$ plane might represent the probabilities assigned for moving one square north from each of the squares on the board. Illegal moves are filtered out before the probability distribution over all $4672$ possible moves is computed.

The body of the network consists of 19 residual blocks \citep{he_residual_2016} made up of rectified covolutional layers and skip connections. The network body leads into two "heads": a value head which produces the scalar evaluation, and a policy head, which produces a stack of planes as detailed above. See \citet{silver_general_2018} for further details.

\subsection{AlphaZero's successor: LCZero}
DeepMind did not open source AlphaZero. The LCZero project was thus born as an attempt to reproduce the work via crowd computing. Though LCZero predominantly uses the same search and evaluation techniques as AlphaZero, the team has made a few improvements. For example, the network architecture adds Squeeze-and-Excitation layers \citep{hu_squeeze-and-excitation_2018} to the residual blocks, and the engine supports endgame tablebases. LCZero has far surpassed the original strength of AlphaZero due to its additional training and improvements.

\section{Chess study: Plaskett's Puzzle}

\subsection{Stockfish 14 performance}
We gain insight into how Stockfish 14 understands Plaskett's Puzzle by inspecting the action-value function $Q$ for the top moves. \medskip
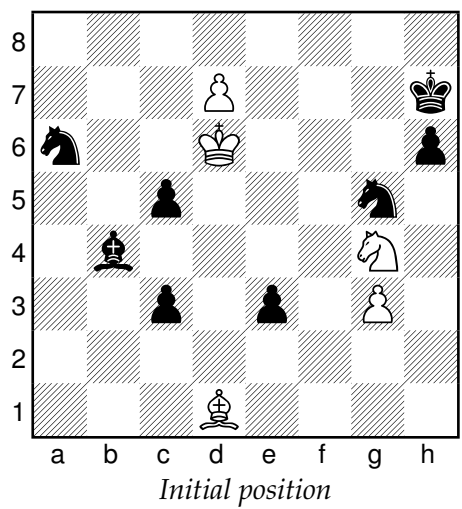
\begin{center}
\fenboard{8/3P3k/n2K3p/2p3n1/1b4N1/2p1p1P1/8/3B4 w - - 0 1}
\showboard \\
\emph{Initial position}
\end{center}

\begin{table}[H]
\centering
\begin{tabular}{@{}llllll@{}}
\toprule
                         & \textbf{d8\symqueen} & \textbf{d8\symrook} & \textbf{\symknight xe3} & \textbf{\symbishop c2+} & \textbf{\symbishop b3} \\ \midrule
\textbf{Q-value}         & -2.84         & -2.92         & -3.16         & -4.19         & -4.40        \\ \midrule
\textbf{Win Probability} & 16.3\%        & 15.7\%        & 14.0\%        & 8.23\%        & 7.36\%       \\ \bottomrule
\end{tabular}
\caption{Stockfish 14 evaluations at depth 30 for the top 5 moves (MultiPV=5) and corresponding approximate win probabilities (calculated according to \citet{isenberg_pawn_2021}). Time on 2.8 GHz Intel Core i5 processor: 0:01:17.}
\label{tab:plaskett-q1}
\end{table}

Given the initial position, Stockfish prefers black, reporting a pawn advantage of -2.84. It does not identify the winning move in its top five choices. However, the engine's evaluation of the position reverses once it analyzes to a greater depth.

\begin{table}[H]
\centering
\begin{tabular}{@{}llllll@{}}
\toprule
                         & \textbf{\symknight f6+} & \textbf{\symknight xe3} & \textbf{\symbishop c2+} & \textbf{d8\symqueen} & \textbf{d8\symrook} \\ \midrule
\textbf{Q-value}         & +3.99         & -3.16         & -4.49         & -4.88         & -4.88        \\ \midrule
\textbf{Win Probability} & 90.9\%        & 14.0\%        & 7.01\%        & 5.68\%        & 5.68\%       \\ \bottomrule
\end{tabular}
\caption{Stockfish 14 evaluations at depth 39 for the top 5 moves (MultiPV=5) and corresponding approximate win probabilities. Time on 2.8 GHz Intel Core i5 processor: 0:18:06.}
\label{tab:plaskett-q2}
\end{table} 

\noindent Van Breukelen originally designed the solution of the puzzle to begin with \mainline{1.Nf6+ Kg7 2.Nh5+ Kg6 3.Bc2+! Kxh5 4.d8=Q}.

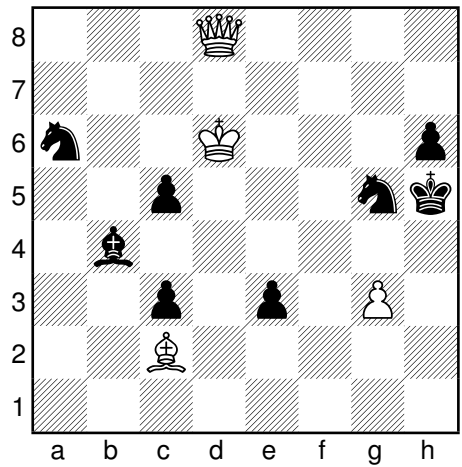
\begin{center}
\showboard \\
\end{center}
The principal variation identified by Stockfish exploits the flaw in the original study. The engine correctly calculates that black can avoid walking into Van Breuklen's forced checkmate with \variation{4...Kg4! 5. Qf6 Kxg3 6. Qe5+ Kf2 7. Qh2+ Ke1 8. Qg1+ Ke2 9. Qg4+ Kf2}. If \textbf{\symknight f7+} is played instead, it leads to a forced checkmate via \mainline{4...Nf7+ 5. Ke6 Nxd8+ 6.Kf5 e2 7.Be4 e1=N! 8.Bd5! c2 9.Bc4 c1=N! 10.Bb5 Nc7 11.Ba4! Ne2 12.Bd1 Nf3 13.Bxe2 Nce6 14.Bxf3#}.

Testing Stockfish against the corrected puzzle (where the knight is moved to h8), we obtain similar results. Though the forced checkmate is 30 half-moves deep, the chess engine reaches a nominal depth of 37 before it detects the move.

\begin{table}[H]
\centering
\begin{tabular}{@{}llllll@{}}
\toprule
                         & \textbf{\symknight f6+} & \textbf{\symknight xe3} & \textbf{d8\symqueen} & \textbf{d8\symrook} & \textbf{\symking c6} \\ \midrule
    \textbf{Q-value}         & $\infty$ (\#15)         & -2.97         & -3.47         & -3.47         & -4.15        \\ \midrule
\textbf{Win Probability} & 100\%        & 15.3\%        & 11.9\%        & 11.9\%        & 8.40\%       \\ \bottomrule
\end{tabular}
\caption{Stockfish 14 evaluations at depth 37 for the top 5 moves (MultiPV=5) and corresponding approximate win probabilities. Time on 2.8 GHz Intel Core i5 processor: 0:24:24.}
\label{tab:plaskett-q3}
\end{table}

For both the original and corrected puzzle, Stockfish's delay in finding the move is due to the heuristics that the engine uses for focusing on more promising branches. For example, it is likely that the knight sacrifice with \variation{3. Bc2+} is sorted near the end of the move list, since alternative moves leave White with more material. Therefore, late move reductions will cause the engine to search variations following \variation{3. Bc2+} to less depth than nominal.

\subsection{LCZero performance}
We test LCZero on the corrected puzzle and find that, even with an extensive amount of computational resources, it does not find the forced checkmate.

\begin{table}[H]
\centering
\begin{tabular}{@{}llllll@{}}
\toprule
                         & \textbf{d8\symrook} & \textbf{\symking c6} & \textbf{d8\symqueen} & \textbf{\symknight f6+} & \textbf{\symbishop c2+} \\ \midrule
\textbf{Q-value}         & -4.67         & -5.01         & -5.48         & -5.60         & -6.47        \\ \midrule
\textbf{Win Probability} & 5.9\%         & 5.5\%         & 5.0\%         & 4.9\%         & 4.2\%        \\ \bottomrule
\end{tabular}
\caption{LCZero evaluations for the top 5 moves after analyzing 60 million nodes.}
\label{tab:plaskett-q4}
\end{table}

\fenboard{7n/3P3k/n2K3p/2p5/1b4N1/2p1p1P1/8/3B4 w - - 0 1}

The engine reports that the best move for black is \variation{1. d8=R} with a pawn advantage of -4.67. It further assigns the winning move, \variation{1. Nf6+}, a pawn advantage of -5.60 and calculates the full continuation as \mainline{1. Nf6+ Kg7 2. Nh5+ Kg6 3. d8=N Kf5 4. Nf4 Ke4 5. Nc6 Nf7+ 6. Ke6 Ng5+ 7. Kf6 Nc7 8. Ne7 Nge6 9. Bc2+ Kf3 10. Bd1+ Kxg3 11. Nd3 Nd4 12. Ke5 Ba3 13. Nf4 c2}.

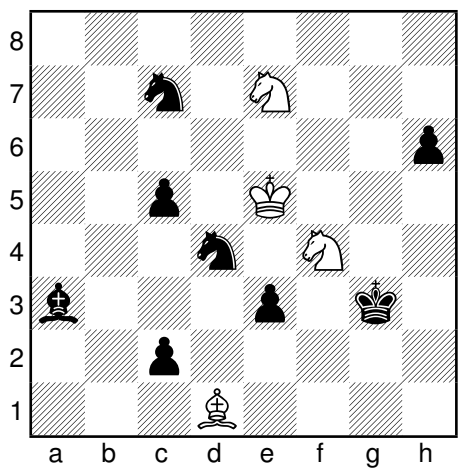
\begin{center}
\showboard \\
\end{center}

The ending position is clearly winning for black.

% additionally tested at cpuct=3.2; still not enough exploration
It is possible to understand Leela's selective search strategy by examining the distribution of positions searched per move. An astounding 92.5\% of the 60 million searched positions follow from \variation{1. Nxe3}. The engine spends only 7.4\% of the time searching positions following \variation{1. Nf6+}, seeing it as less promising. This is partially due to the prior probabilities determined by the policy head. The policy indicates that there is a 15.8\% probability \variation{1. Nxe3} is the optimal move, compared to a 7.4\% probability for \variation{1. Nf6+}. However, the skewed search is also due to subleties in the puzzle. The engine must see the entire checkmate before it is able to realize the benefits, especially regarding the \variation{3. Bc2+} sacrifice made in a materially losing position. The engine thus chooses to prioritize searching other lines instead.

LCZero only finds the forced checkmate after it is given the first move. This happens after searching 5.5M nodes. In the same position, Stockfish searches around 500M nodes before finding the checkmate. This demonstrates LCZero's strength: even if Stockfish searches significantly faster than LCZero, LCZero can find the optimal line of play after searching orders of magnitude less nodes.

\subsection{Fairness of engine comparison}
Historically, engines have been compared by running them on the same hardware. However, the advent of GPU engines has meant that this is not always possible anymore. CPU engines run sub-optimally on GPUs, and GPU engines run sub-optimally on CPUs. The closest we can do is establish rough equivalencies between GPU and CPU hardware.

The "Leela Ratio" factor is based on the ratio of the GPU and CPU evaluation speeds reported in \citet{silver_general_2018}.\footnote{\url{https://github.com/dkappe/leela-ratio}} Its use assumes that the Stockfish vs AlphaZero games ran on "equivalent" hardware. By comparing the GPU and CPU evaluation speeds on our hardware to the Leela Ratio, it is possible to determine whether we give an edge to either engine. We report and recompute an updated Leela Ratio based on the average engine speeds (in nodes per second) from the TCEC Cup 8 Final held in March 2021. Our comparison thus assumes that the TCEC chess engine tournament picks "fair" hardware when it pits CPU and GPU engines against one another.

\[
    F = \frac{\text{Stockfish nps}}{\text{LCZero nps}}
      = \frac{1.5 \times 10^8}{1.4 \times 10^5}
      \approx 1084
\]

This means that during TCEC, Stockfish is allowed to search, on average, \emph{three orders of magnitude} more nodes than LCZero in "fair play." We calculate the Leela Ratio for our experiments as 

\[
    R = F\times\frac{\text{LCZero nodes}}{\text{Stockfish nodes}} = F\times\frac{6.0 \times 10^7}{1.897 \times 10^9} \approx 34
\]

This indicates that we gave LCZero 34 times more computational power than Stockfish. Our experiments show that even with this advantage (in the form of extra time), LCZero does not find the solution to the corrected Plaskett's Puzzle.

\section{Discussion}\label{sec:discussion}

Stockfish and LCZero represent two competing paradigms in the race to build the best chess engine. The magic of the Stockfish engine is programmed into its search; the magic of LCZero into its evaluation. When tasked with solving Plaskett's Puzzle, Stockfish's approach proved superior. The engine searched through nearly 1.9 billion different positions to identify the minimax solution. The algorithm's sheer efficiency—due in part to domain-specific search optimizations—enabled it to find the surprising, unlikely solution. On the other hand, LCZero's selective search was less efficient, primarily because it chose the wrong lines to search deeply. Even its more powerful, deep learning based evaluation function failed to recognize the positional potential of the knight sacrifice. This is particularly notable because LCZero's evaluation function is often unafraid of giving away material for a positional advantage. 

After annotating 40 games between Stockfish and LCZero, FIDE Master Bill Jordan concluded that "Stockfish represents calculation" and "Leela represents intuition" \citep{jordan_calculation_2020}. Like human players, LCZero's intitution allows it to hone in on the most compelling lines. In the majority of positions, this intuition is strong enough to make up for fewer calculations. However, LCZero's human-like approach can fail in edge cases; it is difficult to build a universally-accurate pattern matcher. The safest approach to engine-building may still involve cold, hard calculation, even in seemingly unpromising variations.

That said, there is another aspect of LCZero's algorithm that ought to be considered. The engine was built from "zero" knowledge of the game, other than the rules. AlphaZero's successor, MuZero \citep{schrittwieser_mastering_2020}, removed the dependence on rules altogether \citep{feynman_pleasure_1981}. It stands to reason that with work, LCZero could be trained with this method as well. Such an approach stands in sharp contrast to the large quantity of chess-specific search heuristics required by Stockfish. AlphaZero additionally achieves high levels of performance in Shogi and Go, which demonstrates that the LCZero algorithm is significantly more \emph{generalizable} than Stockfish. In games where humans do not have the knowledge or time to provide domain-specific information, a strong engine can still be trained.

Stockfish and LCZero are examples of what philosopher John Searle calls weak AI \citep{searle_minds_1980}. Weak AI is a tool that performs a task that the mind can perform. This is in contrast to strong AI, which singularly possesses the capabilities of the human mind. Strong AI \emph{understands} the world in the same way that humans do and is often known as artificial general intelligence, or AGI.

Since LCZero learns to play chess without additional human knowledge, its approach demonstrates more potential in the field of AGI. However, the engine's inability to solve Plaskett's Puzzle (given a reasonable amount of resources) means that even the strongest versions of weak AI have not been able to replicate expert human performance. Therefore, we propose two main courses of action for improving LCZero's efficiency on the problem.

One potential solution is additional training. With more self-play, it is likely that LCZero's performance on the puzzle would improve. It may further be possible to train the engine on forced checkmates, which would develop its pattern recognition abilities in mating positions. Such an approach seeks to refine the output of LCZero's policy head through data augmentation \citep{shorten_survey_2019}, improving the engine's intution. Unlike training on expert games, this form of training is impervious to human bias, since the optimal line of play is already known.

Our second proposal involves algorithmic and architectural improvement. Improvements to the search, for example, have already begun with Ceres, which selects lines more intelligently \citep{lczero_team_announcing_2021}. Speed ups might also be gained by simplifying the neural network architecture through different forms of model compression. Common methods include knowledge distillation \citep{bucilua_model_2006, hinton_distilling_2015}, neuron pruning \citep{lecun_optimal_1990, han_learning_2015}, and quantization \citep{jacob_quantization_2018}. With a simplified evaluation function, LCZero can search through more nodes, increasing the likelihood that it will find the optimal line of play.

For a chess problem specific algorithmic improvement, we draw inspiration from how humans solve chess problems. The human approach may not always be the most effective one \citep{campbell_richard_2012}, but it provides us with a starting off point. We notice that humans solve chess problems not only with calculation but also with \emph{imagination}. In the case of Plaskett's Puzzle, it is possible that Tal saw the checkmate in his mind \emph{first} and then found the variation that led him there. We ask whether machines can possess a similar imagination \citep{mahadevan_imagination_2018}.

Since all puzzles have a guaranteed solution, it may be possible for a learner to explicitly predict likely checkmate positions and use these positions to condition the search. Such a process would augment the sequential search process with a deeper, more speculative lookahead. In statistical terms, this may be thought of as "extending the conversation," since we are extending our assessment of win probability by conditioning on the event of reaching a certain checkmate position \citep{lindley_understanding_2013}. It is important to note that in order to verify a checkmate as a forced win, all relevant lines must still be examined. The opponent must truly have no alternative. Humans often optimistically bias their calculations based on their memory of similar positions. They may fall into the trap of "magical thinking," which "involves our inclination to seek and interpret connections between the events around us together with our disinclination to revise belief after further observation" \citep{diaconis_theories_2006}. When implementing a computational form of imagination, it is important to ensure machines do not make the same mistake.

To make progress in AGI, however, systems must do more than improve on a particular task; they must begin to generalize better to new tasks. \citet{silver_general_2018} took important an important step in this direction by building a singular architecture that achieves state-of-the-art results in three different games. Such work begs the question of what a singular model can achieve. Thus, a compelling avenue of research resides in testing changes in performance when AlphaZero learns to play multiple games at once. This can be achieved via multi-task learning \citep{caruana_multitask_1997}, which has seen successes in both NLP \citep{raffel_exploring_2020} and computer vision \citep{zhang_facial_2014}.

For end users, our work implies that Stockfish may currently be the better tool for studying deep puzzles. To confirm this hypothesis, a chess studies dataset ought to be created and used as a benchmark for modern chess engines. For the AI community, our work suggests that even extremely advanced incarnations of weak AI do not outperform humans in all cases. More research will be critical to improving these systems, particularly as we begin to approach the larger goal of building strong AI.

\medskip

\nocite{*}
\setlength\bibitemsep{0.5\baselineskip}
\printbibliography

\end{document}